\documentclass[letterpaper, 10 pt, conference]{ieeeconf}  % Comment this line out if you need a4paper

\IEEEoverridecommandlockouts                              % This command is only needed if 
\usepackage{amsthm}
\usepackage{graphicx}
\usepackage{caption}
\usepackage{subcaption}
\usepackage[linesnumbered,ruled,vlined]{algorithm2e}
\usepackage{algorithmicx}
\usepackage{amsmath}
\theoremstyle{plain}
\usepackage{mathtools}
\usepackage{epsfig}
\usepackage{amssymb}
\usepackage{epstopdf}
\usepackage{multicol}
\usepackage{multirow}
\usepackage{array}
\usepackage{url}
\usepackage{color}
\usepackage{float}
\usepackage{wrapfig}
\usepackage{placeins}
\usepackage{booktabs}

\usepackage{newclude}

\usepackage[english]{babel}

\DeclareMathOperator*{\argmin}{arg\,min}

\title{\LARGE \bf
MPC-MPNet: Model-Predictive Motion Planning Networks for Fast, Near-Optimal Planning under Kinodynamic Constraints
}

\author{Linjun Li$^{*}$, Yinglong Miao$^{*}$, Ahmed H. Qureshi, and Michael C. Yip% <-this % stops a space
\thanks{$^{*}$Equal first author contribution.}
\thanks{L. Li, Y. Miao, A.H. Qureshi, and M.C. Yip are affliated with University of California San Diego, La Jolla, CA 92093, USA.
        {\tt\small \{lili, y2miao, a1qureshi, yip\}@ucsd.edu}}%
}

\begin{document}

\maketitle
\thispagestyle{empty}
\pagestyle{empty}

%%%%%%%%%%%%%%%%%%%%%%%%%%%%%%%%%%%%%%%%%%%%%%%%%%%%%%%%%%%%%%%%%%%%%%%%%%%%%%%%
\begin{abstract}
Kinodynamic Motion Planning (KMP) is to find a robot motion subject to concurrent kinematics and dynamics constraints.  To date, quite a few methods solve KMP problems and those that exist struggle to find near-optimal solutions and exhibit high computational complexity as the planning space dimensionality increases. To address these challenges, we present a scalable, imitation learning-based, Model-Predictive Motion Planning Networks framework that quickly finds near-optimal path solutions with worst-case theoretical guarantees under kinodynamic constraints for practical underactuated systems. Our framework introduces two algorithms built on a neural generator, discriminator, and a parallelizable Model Predictive Controller (MPC). The generator outputs various informed states towards the given target, and the discriminator selects the best possible subset from them for the extension. The MPC locally connects the selected informed states while satisfying the given constraints leading to feasible, near-optimal solutions. We evaluate our algorithms on a range of cluttered, kinodynamically constrained, and underactuated planning problems with results indicating significant improvements in computation times, path qualities, and success rates over existing methods. 

\end{abstract}

%%%%%%%%%%%%%%%%%%%%%%%%%%%%%%%%%%%%%%%%%%%%%%%%%%%%%%%%%%%%%%%%%%%%%%%%%%%%%%%%

\section{INTRODUCTION}
% First Paragraph- Motion Planning and  Kinodynamic Motion Planning, their applications in general, and challenges.
% Motion Planning Background. Introduction to Kinodynamic Motion Planning (Geometry -> Kino constraint).
The motion planning problem is to find a path connecting the given start and goal states while satisfying all the desired constraints on the robot motion. Geometric motion planning is a simple instance that considers only collision-avoidance constraints with states representing the robot's kinematics variables (e.g., position, joint-angles) \cite{lavalle2006planning}. However, in KMP, the state comprises position, velocity, and sometimes acceleration, and the robot motions are required to satisfy both kinematic (e.g., collision-avoidance) and dynamics (e.g., velocity and acceleration) constraints, which makes the problem PSPACE-hard and computationally demanding \cite{lavalle2006planning}. 

The KMP has a broad range of applications from torque-constrained robot manipulation to speed racing \cite{arab2016motion} and performing acrobatic motions \cite{kaufmann2020deep}. The existing state-of-the-art methods solve KMP through Sampling-based Motion Planners (SMPs) \cite{lavalle2006planning} by constructing a tree that originates from a start state and expands by searching the entire state-space to reach the given goal state. The edges of the tree, connecting any two intermediate states, are formed by a local steering function. The local steering often requires solving a boundary-value problem (BVP) through trajectory optimization, which is known to be NP-hard \cite{kacewicz2002complexity} and fails when the boundary constraints are not satisfied. A recent development led to another sampling-based KMP approach \cite{li2016asymptotically}, called Stable Sparse-RRT (SST), that circumvents solving BVPs by using a random shooting method for steering and achieves asymptotic optimality. However, it still takes long computation times from tens of seconds to minutes as environments become more complicated due to uniform exploration of state and control spaces. 

In this paper, we propose an imitation learning-based approach named Model-Predictive Motion Planning Networks (MPC-MPNet)\footnote{Supplementary videos and code release details are available at https://sites.google.com/view/mpc-mpnet}. It has a deep neural networks-based generator and discriminator, which, once trained using expert data, outputs feasible paths that satisfy the given kinodynamic constraints.
%It is a deep neural networks-based, generator-discriminator approach that outputs end-to-end paths while satisfying the given kinodynamic constraints. 
Our approach extends the Motion Planning Networks (MPNet) \cite{qureshi2019motion} and leverages collision-aware Model Predictive Control (MPC) methods in the planning loop for parallelizable local steering. Note that the original MPNet framework \cite{qureshi2019motion} considered only geometric planning problems and relied on bidirectional path expansions and re-planning steps for finding path solutions to given problems. In KMP, the bidirectional path extensions and re-planning steps to repair in-collision path segments often lead to discontinuity and infeasible paths. Therefore, in MPC-MPNet, we propose novel planning algorithms that perform only forward path expansion and avoid re-planning by building a set of informed possible paths. We evaluate our approach in challenging under-actuated robotics problems in complex, cluttered environments. Our results show that MPC-MPNet outperforms state-of-the-art planning algorithms in computational speed and generalizes to unseen planning problems with high success rates.

\section{Related Work}

\begin{figure*}[t!]
\setlength{\belowcaptionskip}{-0.5cm}
    \centering
     \includegraphics[width=0.99\textwidth]{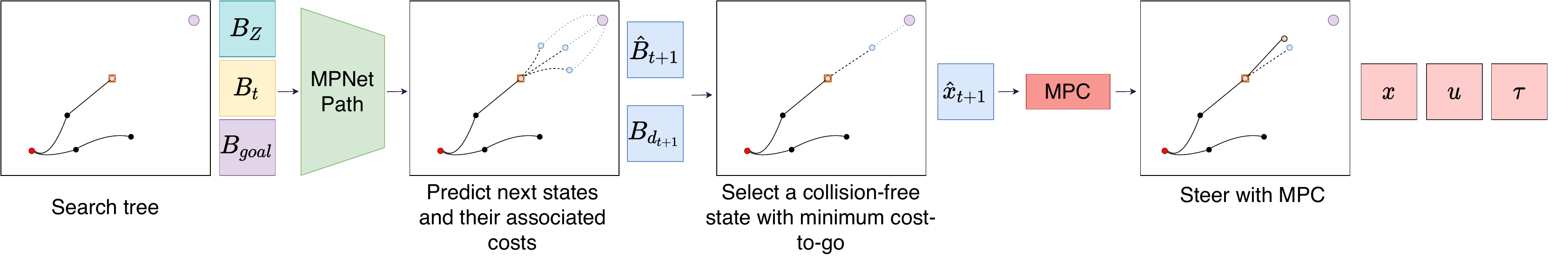}
    \caption{\textbf{MPC-MPNetPath} In each iteration, the neural generator predicts a batch of next states from a given current and select a collision-free state with a minimum estimated cost for the tree expansion using MPC.}
    \label{fig:mpc-mpnetpath}
\end{figure*}

This section presents a brief overview of related work done in the past for solving KMP problems. KMP algorithms depend on the local steering function to satisfy the motion constraints between any two given states. The local steering methods are usually implemented either as (i) random shooting methods which uniformly samples control sequences \cite{donald1993kinodynamic}\cite{lavalle2001randomized}\cite{li2016asymptotically}, (ii) predefined
motion primitives \cite{sakcak2019sampling}, or (iii) local trajectory optimization solvers \cite{xie2015toward}\cite{perez2012lqr}\cite{goretkin2013optimal}\cite{tedrake2010lqr}.

Sampling random control sequences for shooting method has the advantage of worst-case theoretical guarantees. For instance, the SST approach \cite{li2016asymptotically} leverages random exploration of control and state spaces to provide the worst-case asymptotic completeness and optimality guarantees. However, these methods struggle in higher-dimensional and cluttered planning spaces since uniform exploration takes large computation times to determine a feasible path solution. In contrast, the motion primitives methods generate a local database of optimal controllers in their offline stage \cite{sakcak2019sampling} and use them for steering under KMP algorithms. However, as the motion primitive set is finite, it usually does not capture the entire control space and lacks completeness guarantees. 

The trajectory optimization methods formulate local steering as an optimization problem and iteratively solve them for computing action sequences connecting the given states. For instance, \cite{webb2013kinodynamic}\cite{goretkin2013optimal} locally linearize the system and obtain Linear Quadratic Regulators' (LQR) parameters using optimization for steering. Tedrake et. el \cite{tedrake2010lqr} extend LQR methods to construct a tree connecting states within stability regions defined by LQR funnels. However, these methods add a substantial computational burden when constructing trees and do not apply to online planning problems. In a similar vein, Xie et. el \cite{xie2015toward} formulated local steering as BVP and used optimization to find their solutions. These BVP solvers operate in conjunction with traditional SMPs \cite{lavalle2006planning} for finding solutions to KMP problems. However, their approach often collapse when the boundary constraints are not satisfied.

Another class of methods in KMP learn local controllers for steering through reinforcement and imitation learning. For instance, \cite{chiang2019rl} uses reinforcement learning (RL) to acquire local policy for steering within SMP methods. Similarly, \cite{ota2020efficient} \cite{ota2020deep} constructs high-level landmarks, which are later connected by local RL policies. However, these methods inherit RL limitations such as requiring exhaustive interactions with their environments for learning. In the imitation learning paradigm, \cite{wolfslag2018rrt} and \cite{allen2019real} use expert demonstrations to learn local policies for connecting given states, but they still rely on classical planners to generate those state sequences.
%Due to research in deep learning, there is a recent emergence of learning-based motion planning algorithms. We give a brief overview of learning methods for two tasks: object representation and trajectory learning.
%Object representation learning aims at learning a good representation that encodes the obstacle shapes.
%We introduce shape learning for point clouds and voxel inputs, as they are easy to obtain in robotic tasks. For deep learning on point clouds, several novel methods include \cite{qi2017pointnet}\cite{jaritz2019multi}. Their methods are based on the observation that the point cloud is an unordered set \cite{qi2017pointnet}. Hence functions such as sum and max, which preserve the unordered nature, need to be used in the neural networks \cite{qi2017pointnet}. However, PointNet generally suffers from memory and time inefficiency. On the contrary, voxel representation groups point clouds into several cells and achieve a more compact representation of objects \cite{zhou2018voxelnet}.

Recent developments overcome the limitations of classical motion planning by incorporating learned planning distributions into their algorithms \cite{qureshi2019motion}\cite{qureshi2019motionb}\cite{ichter2018learning}\cite{ichter2019robot}. These learning-based planners quickly generate feasible motion sequences online for finding path solutions but mostly consider geometric planning problems. Among them, MPNet \cite{qureshi2019motion} \cite{qureshi2019motionb} generates end-to-end paths by exploiting learned distributions in a bidirectional manner throughout its planning and replanning process. MPNet has also been extended to consider kinematic \cite{qureshi2020compnet} and non-holonomic \cite{johnson2020dynamically} constraints. However, these extensions still rely on bidirectional planning, which often leads to discontinuities and infeasible paths in KMP problems.%In our work, we extend MPNet to KMP problems and propose parallelizable local steering methods and cost metrics for quickly finding path solutions in challenging cluttered environments.
%Trajectory learning has two general fields, reinforcement learning (RL) and imitation learning (IL) \cite{osa2018algorithmic}.
%\cite{pan2017agile} and \cite{cheng2019fast} offer excellent discussion about both. Since our work is closely related to IL, we only give a detailed review of IL. IL can generally be achieved by learning motion primitives \cite{schaal2005learning}\cite{paraschos2013probabilistic}\cite{khansari2011learning}, learning sequential models \cite{qureshi2020motion} and directly learning the distribution of expert trajectories \cite{liu2019state}.
%In \cite{qureshi2020motion}, a transition model is learned to solve the motion planning task. This model is iteratively used to generate a trajectory from the start point to the goal point. Our work can be viewed as an extension of \cite{qureshi2020motion} to consider kinodynamic constraints.

\section{Problem Definition}

\begin{figure*}[t!]
\setlength{\belowcaptionskip}{-0.5cm}
    \centering
     \includegraphics[width=0.99\textwidth]{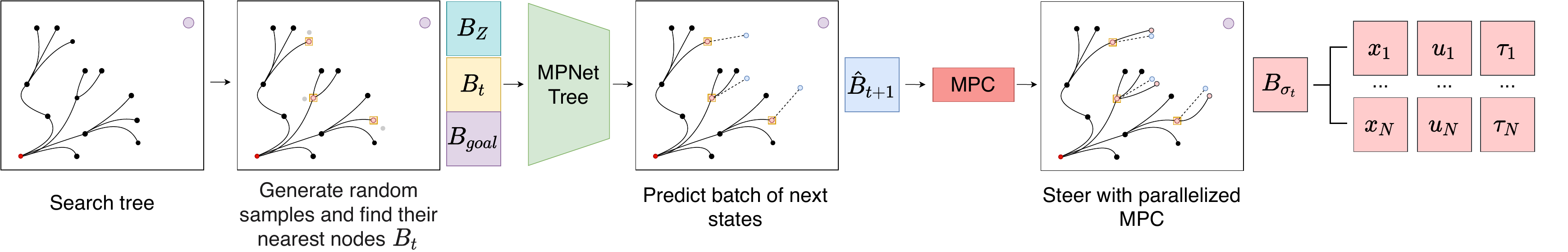}
    \caption{\textbf{MPC-MPNetTree} The neural generator predicts a batch of next states from nearest neighbors of random states inside the search tree. Our parallelized MPC finds the local controllers between randomly selected start and neurally generated next states for extending multiple branches of tree simultaneously towards the given goal state.}
    \label{fig:mpc-mpnettree}
\end{figure*}

Let $\mathcal{C}$ denote a configuration space (C-space) of a mechanical system, where collision and collision-free regions are represented as $\mathcal{C}_{obs}$ and $\mathcal{C}_{free} = \mathcal{C} \backslash \mathcal{C}_{obs}$, respectively. 
Let $\mathcal{X}$ denote the state space in which a state, $\boldsymbol{x}=(\boldsymbol{c}, \dot{\boldsymbol{c}}) \in \mathcal{X}$, contains a configuration $\boldsymbol{c}\in \mathcal{C}$ and the time derivative $\dot{\boldsymbol{c}}$.
%Let $\mathcal{X}$ denotes the state space in which a state, $\boldsymbol{x} \in \mathcal{X}$, comprises a configuration $\boldsymbol{c}\in \mathcal{C}$ and the time derivative $\boldsymbol{\dot{c}}$, i.e., $\boldsymbol{x} = (\boldsymbol{c}, \boldsymbol{\dot{c}})$.
Like C-space, the state space also contains the collision $\mathcal{X}_{obs}$ and collision-free $\mathcal{X}_{free}$ state spaces. %For the given state-space,
The system's dynamics model, represented by an implicit set of equations, can be formulated as $\dot{\boldsymbol{x}} = {f}(\boldsymbol{x},\boldsymbol{u})$, where $\boldsymbol{u}$ denotes the control input to a system from a feasible control set $\mathcal{U}$. In general, the objective of KMP for the given initial state $\boldsymbol{x}_{init}$ and goal region $\mathcal{X}_{goal} \subset \mathcal{X}_{free}$ is to find a collision-free trajectory $\boldsymbol{\sigma}=[(\boldsymbol{x},\boldsymbol{u}, \boldsymbol{\tau})_t]^T_{t=0}$, comprising a sequence of states $[\boldsymbol{x}_t]^T_{t=0}\mapsto \mathcal{X}_{free}$ and controls $[\boldsymbol{u}_t]^T_{t=0}\mapsto \mathcal{U}$ with their corresponding durations $[\boldsymbol{\tau}_t]^T_{t=0}$, such that $\boldsymbol{x}(0)=\boldsymbol{x}_{init}$, $\boldsymbol{x}(T) \in \mathcal{X}_{goal}$.

\section{Model Predictive Motion Planning Networks}
In this section, we present MPC-MPNet, an end-to-end learning-based KMP algorithm. 
MPC-MPNet iteratively generates waypoints and local steering trajectories to construct collision-free paths connecting the start and goal states. It includes a neural generator, discriminator, and two novel planning algorithms named MPC-MPNetPath and MPC-MPNetTree. The main components of our approach are described as follows.

\subsection{Observation Encoder}
The observation encoder embeds the workspace information, represented as voxel maps $\boldsymbol{v}$, into latent features $\boldsymbol{Z}$ containing critical anchor points for the underlying neural generator and discriminator. The voxel maps are volumetric with dimensions $L \times W \times H \times C$, corresponding to length, width, height, and number of channels, respectively, and are usually processed by 3D convolutional neural networks (CNNs). However, we convert the voxel maps into voxel patches with dimensions  $L \times W \times \hat{C}$, where $\hat{C}=HC$, and use the 2D-CNNs for learning the embeddings. This is because 3D-CNNs are known to be computationally inefficient as their representations are inherently cubic and usually contain empty volumes \cite{zhang2018efficient}. 
%The 2D-CNNs are preferred as they are computationally  We do not pick 3D-CNN since it is usually considered inefficient due to sparse inputs.
%as their representations mostly contain empty volumes.

\subsection{Neural Generator}
The neural generator~$G$, with parameters $\boldsymbol{\theta}_g$, is a stochastic neural model that outputs intermediate waypoints $\boldsymbol{\hat{x}}_{t+1}$ for the given environment encoding $\boldsymbol{Z}$, robot's current state $\boldsymbol{x}_t \in \mathcal{X}_{free}$ and goal state $\boldsymbol{x}_{goal} \in \mathcal{X}_{goal}$, i.e.,

\begin{equation}
    \boldsymbol{\hat{x}}_{t+1} \gets G(\boldsymbol{Z}, \boldsymbol{x}_{t}, \boldsymbol{x}_{goal};\boldsymbol{\theta}_g)
    \label{eq:prediction}
\end{equation}
The neural generator adopts Dropout \cite{srivastava2014dropout} in almost every layer to generate stochastic samples. The Dropout randomly selects neurons in each trained network layer, resulting in a sliced model in every forward pass. This operation leads to randomness adapted from expert data. In contrast, other techniques, such as input Gaussian noise, are agnostic of underlying data distribution and incurs difficulty in training deep neural models \cite{creswell2018generative}. 

Our neural generator is trained end-to-end with the observation encoder by optimizing the following mean square error between the predicted states $\boldsymbol{\hat{x}}$ and the demonstration trajectories' states $\boldsymbol{x}^*$, i.e.,
\begin{equation}
    % \theta^o, \theta^g \gets \mathop{\arg\min}_{\theta^e,\theta^p}
    L_{G_{\boldsymbol{\theta}_g}} = \frac{1}{N_{p}}\sum_{i=0}^{N}\frac{1}{T_i}\sum_{j=0}^{T_i-1}
    \|\boldsymbol{\hat{x}}_{i,j+1}-\boldsymbol{x}^*_{i,j+1}\|^2
    \label{eq:train-generator}
\end{equation}
where $N_{p}$ is the total number of paths and $T_i$ is the length of each path $i$ in the dataset.

\subsection{Neural Discriminator}
%Since the neural generator approximates the implicit planning distributions, some predictions might be anomalous. 
Due to stochasticity in the neural generator, the predictions are usually scattered towards the given target. To select the best state from the given set, we introduce a neural discriminator network that predicts a given state's time-to-reach the desired target. Our planning procedures leverage the time-to-reach predictions as a cost to prune outputs of the neural generator.

Hence, the discriminator $D$, with parameters $\boldsymbol{\theta}_d$, takes environment embedding $\boldsymbol{Z}$, the robot state $\boldsymbol{x}_t$ and $\boldsymbol{x}_{goal}$ as input, and predicts the cost as:
\begin{equation}
    \begin{aligned}
        \hat{d}_{t} \gets  D(\boldsymbol{Z}, \boldsymbol{x}_{t}, \boldsymbol{x}_{goal};\boldsymbol{\theta}_d)\\
    \end{aligned}
    \label{eq:prediction-cost}
\end{equation}
The neural discriminator is trained to minimize the mean square error between the predicted costs and the real costs:
\begin{equation}
    % \theta^d \gets \mathop{\arg\min}_{\theta^d} 
     L_{D_{\boldsymbol{\theta}_d}} = \frac{1}{N_p}\sum_{i=0}^{N}\frac{1}{T_i}\sum_{j=0}^{T_i-1}
        \|\hat{d}_{i,j+1}-(\sum_{k=j+1}^{T_i}{d}^*_{i,k})\|^2    \label{eq:train-cost}
\end{equation}
where ${d}^*_{i,k}$ is the cost to goal at waypoint $k$ in the path $i$. 

To balance the positive and negative training samples and enhance the discriminator's performance, we augment the training data by assigning large penalties to in-collision waypoints and unreachable transition pairs. As a result, the trained discriminator predicts high costs for invalid states and eliminates anomalous waypoints.

\subsection{Model Predictive Control}
To satisfy the kinodynamic constraints during tree/path expansion, we utilize MPC as a steering function, a strategy widely used for optimal control problems. We use MPC for several reasons, including implementation simplicity, lower computational complexity, and parallel computation potential. Furthermore, compared to BVP solvers, MPC models the trajectory generation process as an initial-value problem. Therefore, MPC does not collapse when the terminal state is unreachable and instead returns a nearest, valid terminal state. In contrast, BVP solvers fail in such situations because boundary conditions are not satisfied. 

MPC takes a current $\boldsymbol{x}_t$, a target state $\hat{\boldsymbol{x}}_{t+1}$, and generates an optimized trajectory $\boldsymbol{\sigma}_t=[(\boldsymbol{x}, \boldsymbol{u}, \boldsymbol{\tau})_t]$ minimizing the cost between the propagated terminal state and $\hat{\boldsymbol{x}}_{t+1}$. Our MPC solver is implemented with Cross Entropy Method (CEM) \cite{Botev_thecross-entropy} and time-elastic-band approach to optimize both control and their duration sequences \cite{7331052}. CEM takes advantage of the Monte Carlo method and importance sampling to estimate the optimal trajectories distribution iteratively. 

Algorithm \ref{algo:cem-mpc} outlines our MPC approach. Using sampled controls and durations from a parameterized distribution $(\boldsymbol{u}, \boldsymbol{\tau})_i\sim \mathcal{D}(\boldsymbol{u},\boldsymbol{\tau}; \boldsymbol{\theta}_{mpc})$, we propagate and generate the steering trajectory $\boldsymbol{\sigma}_i$, from the given starting state $\boldsymbol{x}_t$.
% The terminal state of the steering trajectory is denoted as $\boldsymbol{\sigma_{local}}\boldsymbol{.x}[-1]$.
These propagations are ranked through a cost function defined as $d_s=d(\boldsymbol{\sigma}_i, \boldsymbol{\hat{x}}_{t+1})+d_c(\boldsymbol{\sigma}_i)$, where $d(\cdot)$ is the distance metrics between given states and $d_c(\cdot)$ is a collision penalty function. The cost function selects the elite samples, i.e., the propagated states with the lowest scores. These elite states are used to update MPC parameters $\boldsymbol{\theta}_{mpc}$ by minimizing the cross-entropy loss between the distributions of steered terminal states and the target state $\boldsymbol{\hat{x}}_{t+1}$. Note that, the collision penalty discourages in-collision trajectory expansions. Furthermore, in our implementation, we assume the distribution $\mathcal{D}(\boldsymbol{u},\boldsymbol{\tau};\boldsymbol{\theta}_{mpc})$ of controls and their durations at each time step  to be Gaussian distributions.

\begin{algorithm}[ht]
initialize parameters $\boldsymbol{\theta}_{mpc}$\;
\For{iter $\gets 1$ \KwTo $N_{iter}$}
{
    \For{$\boldsymbol{u}_i, \boldsymbol{\tau}_i \sim \mathcal{D}(\boldsymbol{u}, \boldsymbol{\tau}; \boldsymbol{\theta}_{mpc})$}
    {
        Propagate $\boldsymbol{x_t}$ with $\boldsymbol{u}_i,\boldsymbol{\tau}_i$ to generate 
        $\boldsymbol{\sigma}_i$\;
        %Evaluate $d_{s_i} = d(\boldsymbol{\sigma}_i, \boldsymbol{\hat{x}}_{t+1})+d_c(\boldsymbol{\sigma}_i)$\;
        Select elite samples based on their scores\;
        Update $\boldsymbol{\theta}_{mpc}$ with elite samples\;
        % Keep best index
        % $k=\argmin_i\{d_{s,\cdot}\}$\, and select
        % ${\boldsymbol{\sigma^*_{local}}}\gets\boldsymbol{\sigma^k_{local}}
        % $\;
        Update optimal trajectory with the best $\boldsymbol{\sigma}^*_i$
    }
}
\Return ${\boldsymbol{\sigma}^*_i}$\;
\caption{Model Predictive Control ($\boldsymbol{x}_t, \hat{\boldsymbol{x}}_{t+1}$)}
\label{algo:cem-mpc}
\end{algorithm}

\subsection{Parallelization}
We use neural networks to process multiple waypoints as a batch to improve performance and implement our MPC algorithm using tensors on Graphic Processing Units (GPUs) for parallel processing.
With parallel computation, the neural networks and MPC are accelerated to process up to $N_B \in \mathbb{N}$ samples simultaneously. To represent batch parallel processing, we introduce new notations denoting all vectorized inputs in batch form using the symbol $\boldsymbol{B}$, i.e., 
%Denote multiple robot states  as $\boldsymbol{B}$ at each time with repeated single robot state or multiple different states $\boldsymbol{x}_t$, and goals as $\boldsymbol{B}_{goal}$:
\begin{equation}
   \boldsymbol{B}_t = 
        \begin{bmatrix}
            \boldsymbol{x}^{1}_t\\
            \boldsymbol{x}^{2}_t\\
            \vdots\\
            \boldsymbol{x}^{N_B}_t
        \end{bmatrix},
    \boldsymbol{B}_{goal} = 
    % \mathrm{Expand}(\boldsymbol{x_{goal}})=
    \begin{bmatrix}
            \boldsymbol{x}_{goal}\\
            \boldsymbol{x}_{goal}\\
            \vdots\\
            \boldsymbol{x}_{goal}
        \end{bmatrix},
    \boldsymbol{B}_{Z}  = 
    % \mathrm{Expand}(Z) =
    \begin{bmatrix}
            \boldsymbol{Z}\\
            \boldsymbol{Z}\\
            \vdots\\
            \boldsymbol{Z}
        \end{bmatrix},
\end{equation}
% \begin{align}
%     B_{x} &= 
%         \begin{bmatrix}
%             \boldsymbol{x_{1}}&
%             \boldsymbol{x_{2}}&
%             \hdots&
%             \boldsymbol{x_{N_B}}
%         \end{bmatrix}^T\\
%     B_{goal} &= 
%     % \mathrm{Expand}(\boldsymbol{x_{goal}})=
%     \begin{bmatrix}
%             \boldsymbol{x_{goal}}&
%             \boldsymbol{x_{goal}}&
%             \hdots&
%             \boldsymbol{x_{goal}}
%         \end{bmatrix}^T\\
%     B_{Z} & = 
%     % \mathrm{Expand}(Z) =
%     \begin{bmatrix}
%             \boldsymbol{Z}&
%             \boldsymbol{Z}&
%             \hdots&
%             \boldsymbol{Z}
%         \end{bmatrix}^T
% \end{align}
where $\boldsymbol{B}_t$, $\boldsymbol{B}_{goal}$, and $\boldsymbol{B}_Z$ correspond to the batch of current states, desired states, and observation encodings. 

During execution, our stochastic generator outputs a variety of next states $\boldsymbol{B}_{t+1}$, and the discriminator predicts their associated costs $\boldsymbol{B}_{d_{t+1}}$ as:
\begin{equation}
     \hat{ \boldsymbol{B}}_{t+1} \gets G( \boldsymbol{B}_{Z},  \boldsymbol{B}_t,  \boldsymbol{B}_{goal};\boldsymbol{\theta}_{g})
     \label{eq:prediction-batch}
\end{equation}
\begin{equation}
    \boldsymbol{B}_{d_{t+1}} \gets D( \boldsymbol{B}_Z, {\hat{ \boldsymbol{B}}_{t+1}},{ \boldsymbol{B}_{goal}};\boldsymbol{\theta}_{d})
    \label{eq:cost-selection}
\end{equation}
%and only the waypoint with minimum estimated cost is select to be steered as $\boldsymbol{\hat{x}}_{new}$.
For the given start $\boldsymbol{B}_{t}$ and target $\hat{\boldsymbol{B}}_{t+1}$ states, our parallelized MPC generates their corresponding local kinodynamic trajectories as:
\begin{equation}
\boldsymbol{B}_{\boldsymbol{\sigma}_t}\gets \textrm{MPC}(\boldsymbol{B}_t, \hat{\boldsymbol{B}}_{t+1})
\label{eq:cemmpc-parallel}
\end{equation}
where $\boldsymbol{B}_{\boldsymbol{\sigma}_t}=[(\boldsymbol{x}_i,\boldsymbol{u}_i,\boldsymbol{\tau}_i)_t]^{N_B}_{i=1}$ is a batch of local trajectories at time $t$.
%More precisely, in CUDA implementation, each transit-pair will be assigned to a separate thread block to manage CUDA threads needed for that sub-problem, and all the sampling and model propagation is processed in different threads parallelly based on threadIdx. 

\begin{algorithm}
${T}\gets\{\boldsymbol{x}_{init}\}$,\:$\boldsymbol{B}_t \gets \boldsymbol{x}_{init}$\;
$\boldsymbol{B}_Z \gets \boldsymbol{Z}$,\:$\boldsymbol{B}_{goal} \gets \boldsymbol{x}_{goal}$\;

\For{$i\gets 1$ \KwTo $n$}
{   
    $\hat{\boldsymbol{B}}_{t+1}
    \gets 
    % \textrm{G}(\boldsymbol{B}_Z, \boldsymbol{B_x}(\boldsymbol{x}), \boldsymbol{B}_{goal})
    G(\boldsymbol{B}_Z, \boldsymbol{B}_t, \boldsymbol{B}_{goal}; \boldsymbol{\theta}_g)
    $\;
    
    $\hat{\boldsymbol{x}}_{t+1}\gets \argmin_{\hat{x}_{t+1}} D({\boldsymbol{B}_Z, \hat{\boldsymbol{B}}_{t+1}, \boldsymbol{B}_{goal}}; \boldsymbol{\theta}_d)$\;
    
    $\boldsymbol{\sigma}_{t} \gets MPC(\boldsymbol{x}_t, \hat{\boldsymbol{x}}_{t+1})$\;

    \If{$\mathrm{Invalid}(\boldsymbol{\sigma}_t)$}
    {
        $\boldsymbol{B}_t \gets \textrm{randomNode}({T})$\;
    }
    \Else
    {
        $\mathrm{addToTree}(\boldsymbol{\sigma}_t,T)$\;
        set batch $\boldsymbol{B}_t$ with a terminal state of $\boldsymbol{\sigma}_t$\;
    }
    % \If{$d(\mathrm{T},s_{goal})\leq V_G$}
    \If {$\mathrm{Reached}({T}, \boldsymbol{x}_{goal})$}
    {
        \Return $\mathrm{ExtractPath}(T)$\;
    }
}
\Return $ \varnothing $\;

\caption{MPC-MPNetPath ($\boldsymbol{Z},\boldsymbol{x}_{init}, \boldsymbol{x}_{goal}$)}
\label{algo:mpc-mpnetpath}
\end{algorithm}

\begin{algorithm}[t!]
$T \gets\{ \boldsymbol{x}_{init} \}$\;
$\boldsymbol{B}_Z \gets \boldsymbol{Z}$,\:
$\boldsymbol{B}_{goal} \gets \boldsymbol{x}_{goal}$\;
\For{$i\gets 1$ \KwTo $n$}
{
    $\boldsymbol{B}_{rand}\gets\mathrm{RandomSample}()$\;
    $\boldsymbol{B}_t\gets\mathrm{NearestNeighbor}(\boldsymbol{B}_{rand}, T)$\;
    
    $\hat{\boldsymbol{B}}_{t+1}\gets G(\boldsymbol{B}_Z, \boldsymbol{B}_t, \boldsymbol{B}_{goal};\boldsymbol{\theta}_{g})$\;
    
    $\boldsymbol{B}_{\boldsymbol{\sigma}_t}\gets MPC(\boldsymbol{B}_t, \hat{\boldsymbol{B}}_{t+1})$\;
    $\mathrm{addToTree}(\boldsymbol{B}_{\boldsymbol{\sigma}_t},T)$\;
    % \If{$d(T,x_G)\leq V_G$}
    \If {$\mathrm{Reached}({T}, \boldsymbol{x}_{goal})$}
    {
        \Return $\mathrm{ExtractPath}(T)$\;
    }
}
\Return $ \varnothing $\;
\caption{MPC-MPNetTree ($\boldsymbol{Z}, \boldsymbol{x}_{init}, \boldsymbol{x}_{goal}$)}
\label{algo:mpc-mpnettree}
\end{algorithm}

\subsection{Planning Algorithms}
In this section, we present our planning algorithms that balance exploration-exploitation in their different ways for quickly finding a path solution with a unidirectional tree expansion.\\\\
\textbf{MPC-MPNetPath:} Figure \ref{fig:mpc-mpnetpath} and Algorithm~\ref{algo:mpc-mpnetpath} outline our MPC-MPNetPath algorithm. The procedure begins by generating a batch of new samples using the neural generator $G$. The discriminator $D$ prunes the generated batch $\boldsymbol{B}_t$ by selecting a sample $\hat{\boldsymbol{x}}_{t+1}$ with a minimum cost $\hat{d}$ to reach the given target $\boldsymbol{x}_{goal}$. The MPC module takes the current node $\boldsymbol{x}_{t}$ and selected next state $\hat{\boldsymbol{x}}_{t+1}$ to perform the kinodynamic steering, leading to a local trajectory $\boldsymbol{\sigma}_t$. The terminal state of $\boldsymbol{\sigma}_t$ is the resulting valid state $\boldsymbol{x}_{t+1}$ (as close as possible to $\hat{\boldsymbol{x}}_{t+1}$ while satisfying constraints) after the execution of $\boldsymbol{u}$ on state $\boldsymbol{x}_{t}$ for time duration $\boldsymbol{\tau}$. The local trajectory is added to the tree if it is valid. In the case of invalid local trajectory, i.e., not collision-free, a random node is selected from the tree and is treated as new current state $\boldsymbol{x}_{t}$ for the next planning iteration. Once, the goal is reached, an end-to-end path is extracted connecting the given start and goal states under kinodynamical constraints.\\\\
\textbf{MPC-MPNetTree:} This method leverages (i) the innate capacity of neural networks to process batches, and (ii) our parallelized MPC framework to expand multiple nodes of the search tree simultaneously, directed towards the given target states $\boldsymbol{B}_{goal}$. The algorithm is summarized in Figure \ref{fig:mpc-mpnettree} and Algorithm \ref{algo:mpc-mpnettree}. In each iteration, MPC-MPNetTree samples a set of random states $\boldsymbol{B}_{rand}$ using the $\mathrm{RandomSample}$ function and finds their corresponding nearest neighbors in the tree by calling the $\mathrm{NearestNeighbor}$ function. These nearest nodes are treated as current set of states $\boldsymbol{B}_t$ for the underling MPC-MPNet procedures, i.e., the generator outputs the next batch of samples and MPC computes their local trajectories $\boldsymbol{B}_{\boldsymbol{\sigma}_t}$. The valid local trajectories are added to the search tree, and the final path is extracted once the tree reaches to the given goal state. 
\begin{figure*}[h]
    \begin{subfigure}[b]{\textwidth}
        \includegraphics[width=\textwidth]{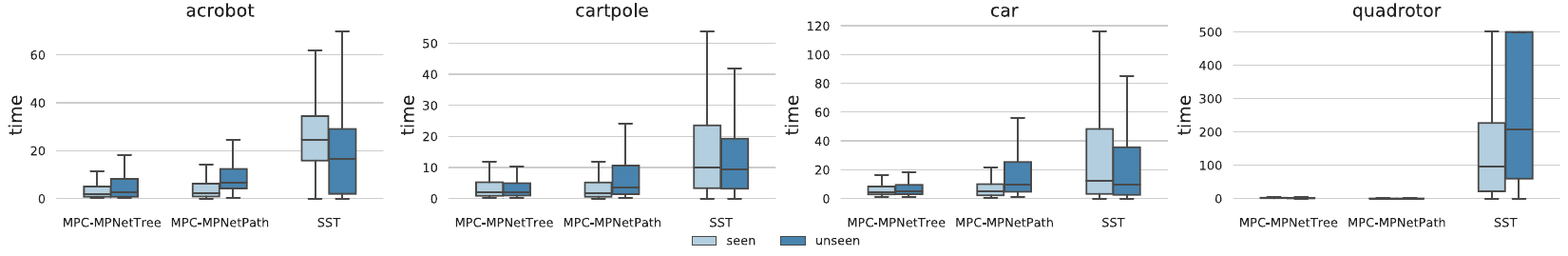}
         \caption{Computational time (in seconds) comparison for Acrobot, Cart-pole, Car and Quadrotor in seen and unseen environments}
         \label{fig:plot-time}
     \end{subfigure}
    \begin{subfigure}[b]{\textwidth}
         \centering
        \includegraphics[width=\textwidth]{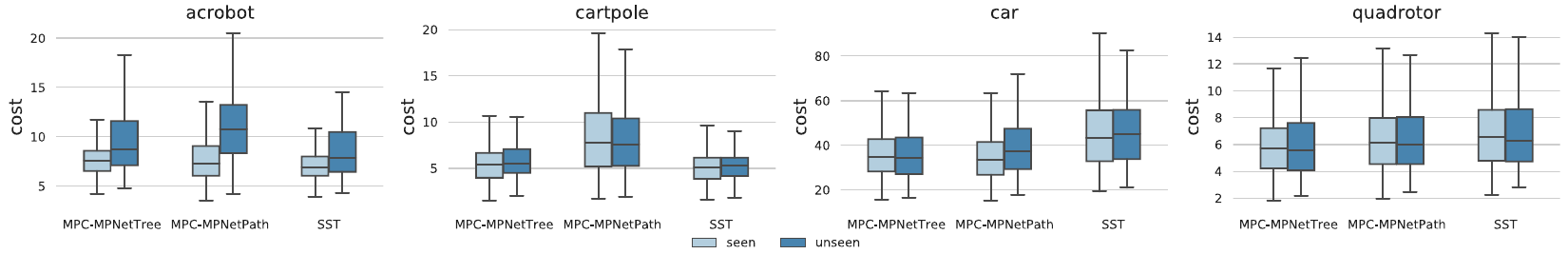}

         \caption{Path cost comparison for Acrobot, Cart-pole, Car and Quadrotor in seen and unseen environments}
         \label{fig:cost-acrobot-cartpole-car}
     \end{subfigure}     
    \caption{The interquartile ranges of computation times and path qualities (time-to-reach the target) for MPC-MPNetPath, MPC-MPNetTree, and SST in Acrobot, Cart-pole, Car and Quadrotor environments.}\vspace*{-0.1in}
    \label{fig:boxplot-acrobot-cartpole-car}
\end{figure*}

%%%%%%%%%%%%%%%%%%%%%
% END of Box Plots
%%%%%%%%%%%%%%%%%%%%%

\begin{figure}[ht!]
    \begin{subfigure}[t]{0.3\linewidth}
        \centering
        \includegraphics[width=\linewidth]{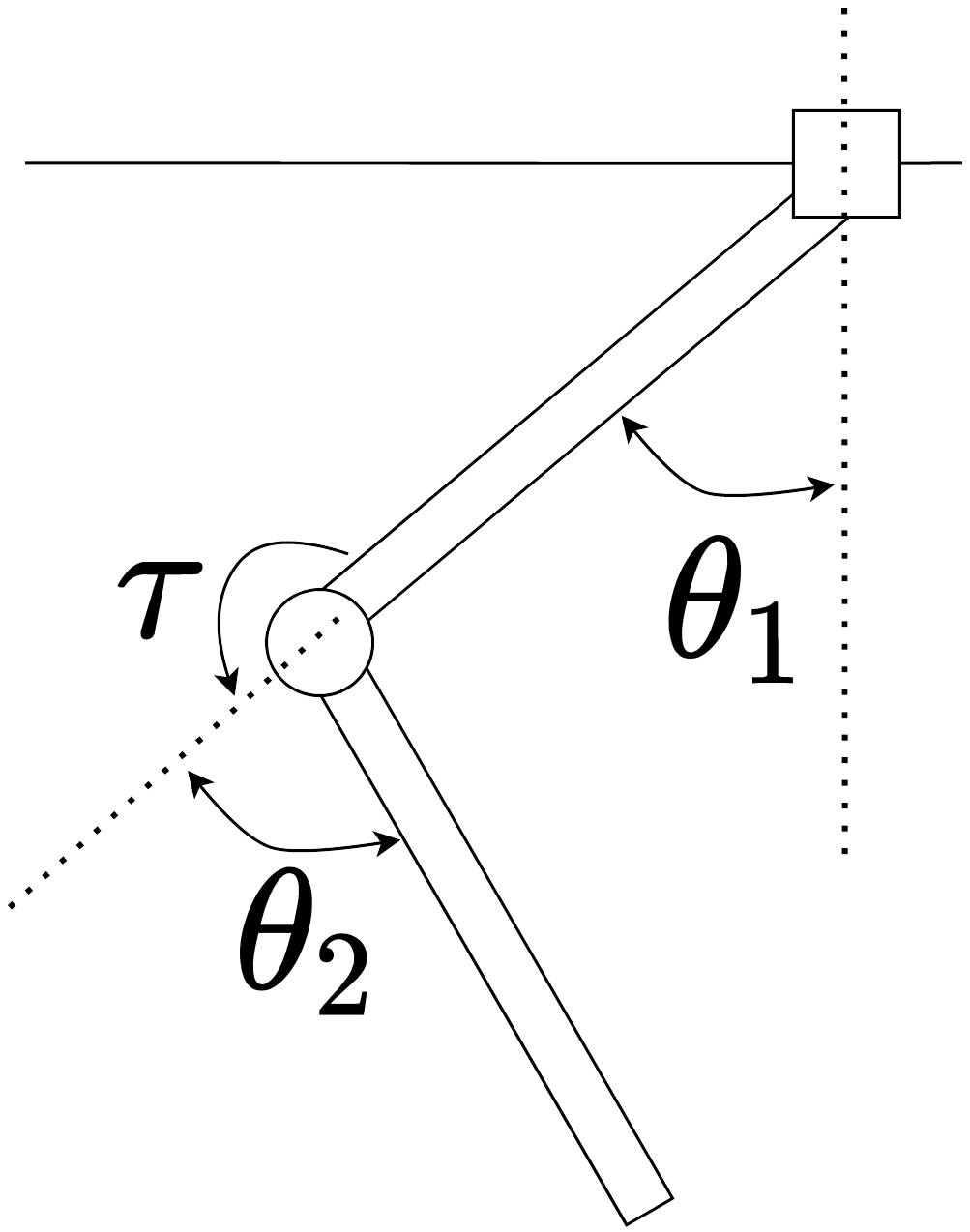}
        \caption{}
    \end{subfigure}
    \begin{subfigure}[t]{0.18\linewidth}
        \centering
        \includegraphics[width=\linewidth]{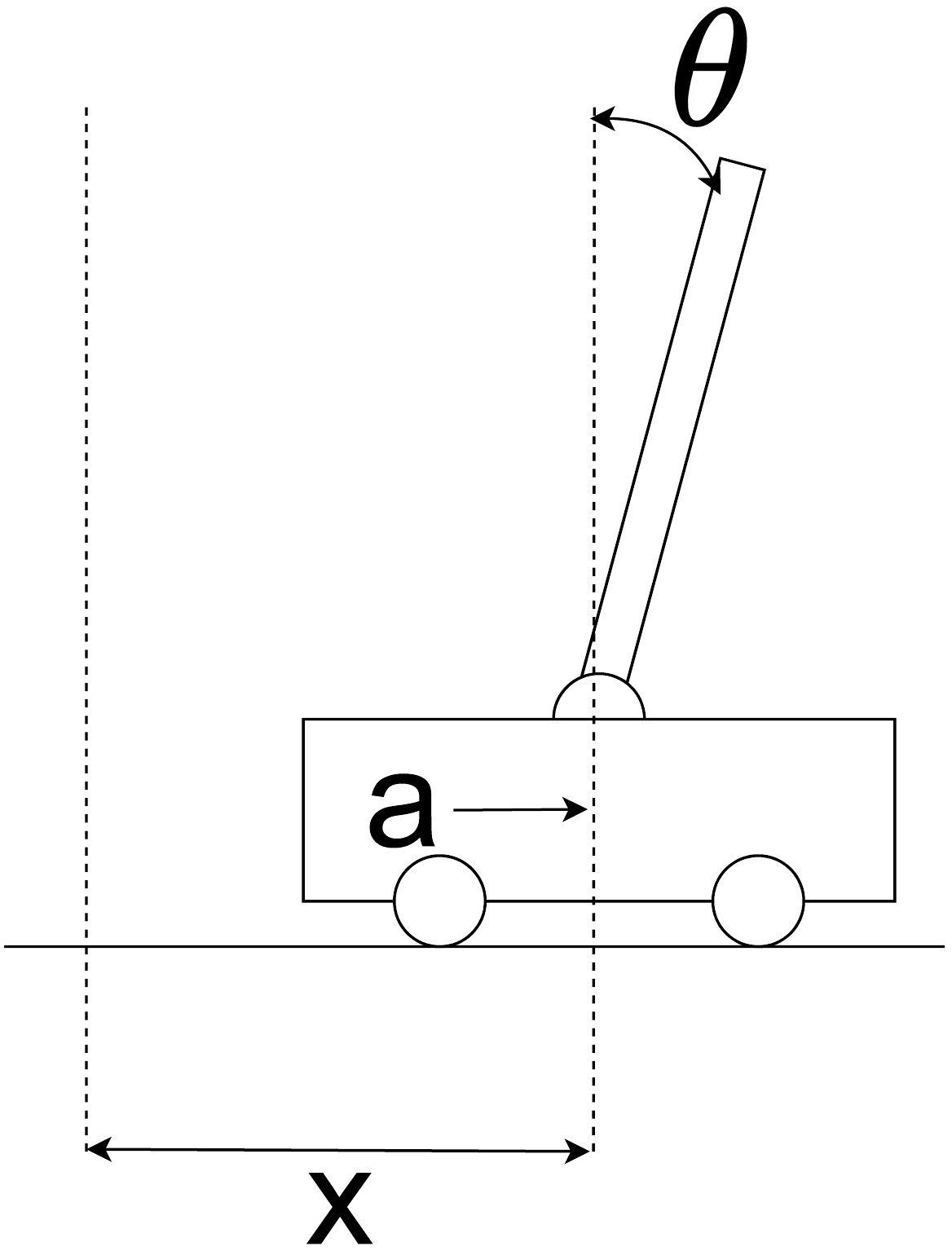}
        \caption{}
    \end{subfigure}
    \begin{subfigure}[t]{.18\linewidth}
        \centering
        \includegraphics[width=\linewidth]{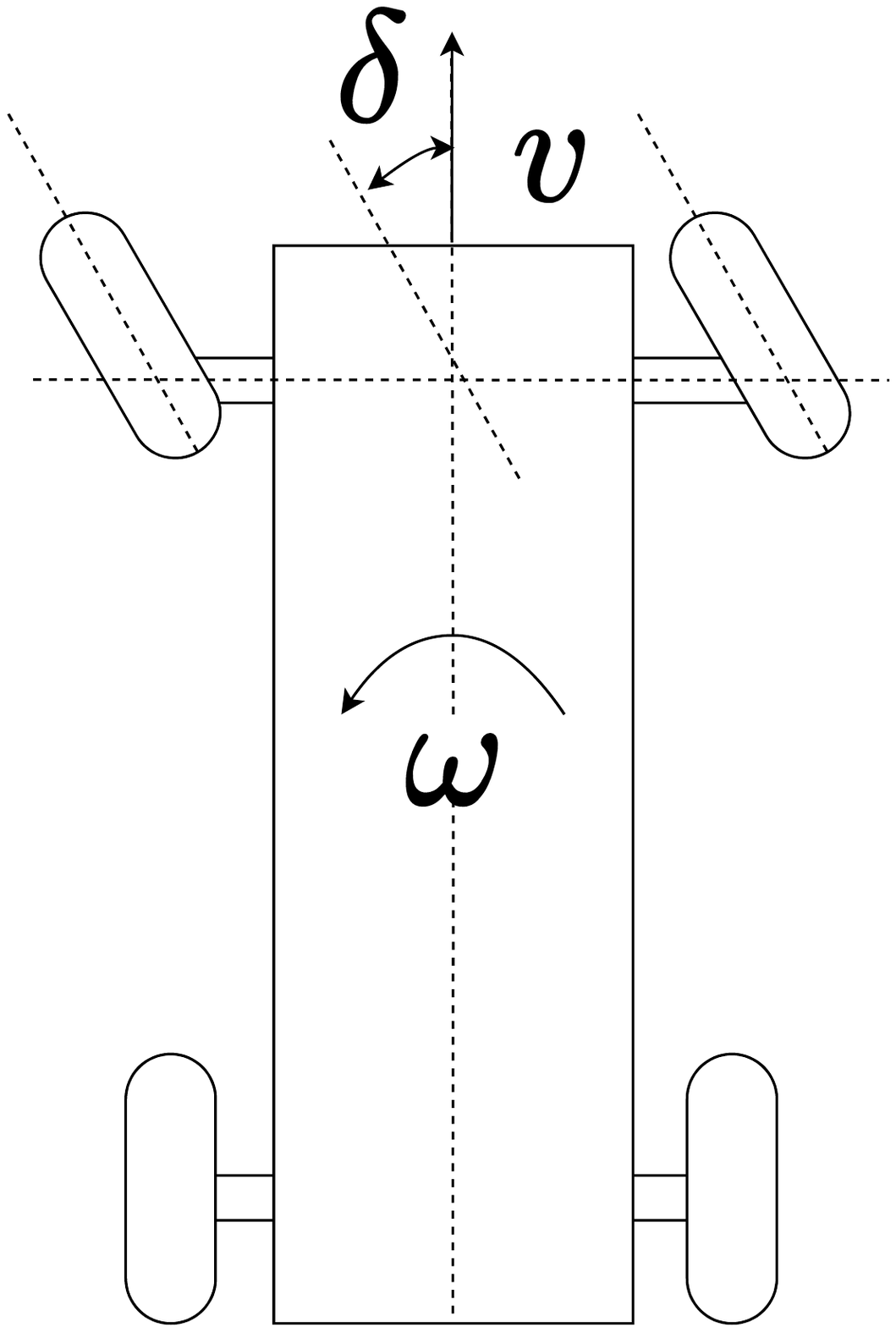}
        \caption{}
    \end{subfigure}
    \begin{subfigure}[t]{.31\linewidth}
        \centering
        \includegraphics[width=\linewidth]{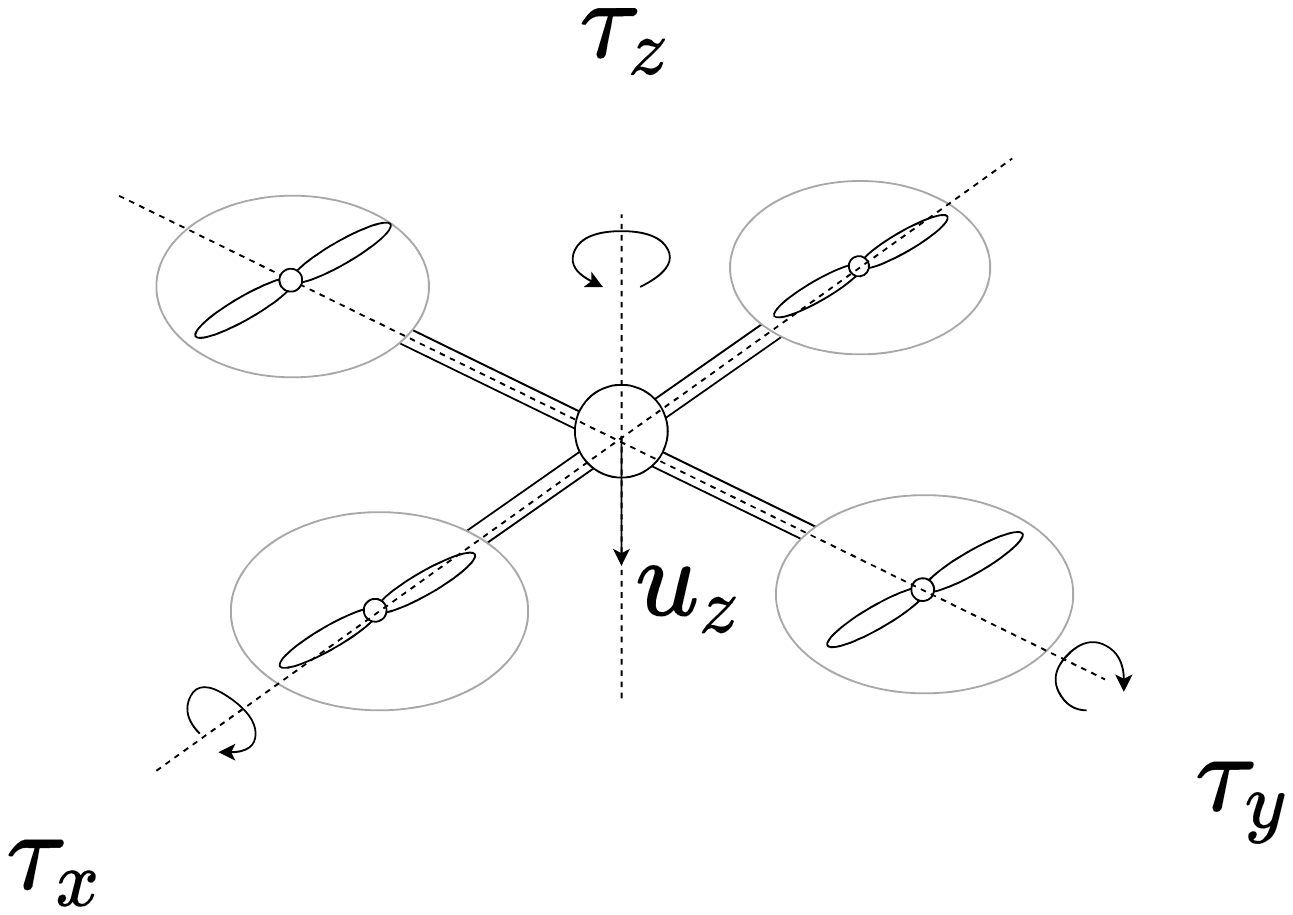}
        \caption{}
    \end{subfigure}
    \caption{We consider the following robotic systems, (a) Acrobot, (b) Cartpole, (c) Car, and (d) Quadrotor, with complex dynamics for our cluttered, kinodynamically constrained environments.}
    \label{fig:schematics}\vspace*{-0.13in}
\end{figure}
\section{Implementation Details}

%%%%%%%%%%%%%%%%%%%%%%%%%%%%%%%%
% Results Table                %
%%%%%%%%%%%%%%%%%%%%%%%%%%%%%%%%

\begin{table*}[bt!]
\centering \scriptsize
\begin{tabular}{ccccc}\toprule
\multirow{2}{*}{Methods}&
\multicolumn{4}{c}{Planning Tasks}\\
\cmidrule{2-5}&
\multicolumn{1}{c}{Acrobot}&
\multicolumn{1}{c}{Cart-Pole}&
\multicolumn{1}{c}{Quadrotor}&
\multicolumn{1}{c}{Car-like}\\
\midrule

%&& &&& &\\ \hline
\multirow{1}{*}{MP-Path}& 
\multirow{1}{*}{$5.09 \pm 6.87$ $(9.88 \pm 8.70)$} &  %$(93.3\%)
\multirow{1}{*}{$4.74 \pm 7.37$ $(7.93 \pm 9.13)$} & %$(91.7\%)$
\multirow{1}{*}{$\boldsymbol{0.46 \pm 2.12}$ $(\boldsymbol{0.41 \pm 1.12})$}& %99.9\%%
\multirow{1}{*}{$8.96 \pm 12.33$ $(19.10 \pm 21.25)$}\\ %99.2\%

\multirow{1}{*}{MP-Tree}&  
\multirow{1}{*}{$\boldsymbol{4.19 \pm 6.03}$ $(\boldsymbol{6.13 \pm 7.90})$}  &% \boldsymbol{96.5\%}
\multirow{1}{*}{$\boldsymbol{4.43 \pm 6.07}$ $(\boldsymbol{4.36 \pm 5.55})$} & % \boldsymbol{97.25}\%
\multirow{1}{*}{$2.26 \pm 3.26$ ${(1.78 \pm 2.45)}$}& %\boldsymbol{100.0}\%
\multirow{1}{*}{$\boldsymbol{7.51 \pm 8.72}$ $(\boldsymbol{8.40 \pm 11.48})$} \\ % \boldsymbol{99.25}\%

\multirow{1}{*}{SST}& 
\multirow{1}{*}{$28.32 \pm 20.53$ $(21.37 \pm 23.02)$} &
\multirow{1}{*}{$14.99 \pm 14.29$ $(12.21 \pm 10.11)$}  &
\multirow{1}{*}{$143.69 \pm 143.43$ $(251.03 \pm 197.83)$}&
\multirow{1}{*}{$41.69 \pm 70.31$ $(28.16 \pm 40.00)$} \\ 

\bottomrule
\end{tabular}
\caption{\small  The total mean computation times with standard deviations in seen and unseen (reported inside parenthesis) test environments are presented for MPC-MPNetPath (MP-Path), MPC-MPNetTree (MP-Tree), and SST in various kinodynamic planning problems.} 
\label{tab:results}
\vspace*{-0.1in}\end{table*}
%%%% tables end here
%%%%%%%%%%%%%%%%%%%%%%%%%%%%%%%%
% Results Table ends               %
%%%%%%%%%%%%%%%%%%%%%%%%%%%%%%%%

This section describes the necessary implementation details of our frameworks with their training and testing environments. We implement our neural modules using Pytorch and export them to C++ with Torchscript. Our parallelized MPC algorithm follows standard GPU programming. For the training and testing environments, we consider the following kinodynamically constrained, cluttered environments with schematics shown in Fig. \ref{fig:schematics}. 
\subsubsection{Acrobot}
We use the acrobot dynamics as specified in \cite{spong1998underactuated}. The state space is defined as $[\theta_1,\theta_2.\dot{\theta_1},\dot{\theta_2}]
\in
[-\pi,\pi]^2\times[-6,6]^2$.
The control space is defined as $[-4,4]$.
We generate four rectangular obstacles and restrict the center of the obstacles to lie inside the annulus that affects the acrobot movement.
\subsubsection{Cartpole}
The cartpole dynamics are used as specified in \cite{papadopoulos2014analysis}. The state space is defined as: $[x,\dot{x},\theta,\dot{\theta}]\in [-30,30]\times[-40,40]\times[-\pi,\pi]\times[-2,2]$. and the control space is defined as $[-300,300]$. We randomly place seven rectangular obstacles in the environment to challenge and restrict the cartpole motion.
\subsubsection{Car}
This is a 2D first-order car system, where the state space is of 3DOF, including position and orientation. The control inputs are the position velocity and angular velocity.  
The state space is defined as
$[x,y,\theta]\in [-25,25]\times[-35,35]\times[-\pi,\pi]$ and control space bound as $[0,2]\times[-0.5,0.5]$. 
We randomly place five rectangular obstacles in the workspace, and limit the distance between obstacles to ensure narrow passages.
\subsubsection{Quadrotor}
We define the quadrotor dynamics as in \cite{ai2013integrated}.
The state space is defined as: $[p,q,\dot{p}, {\omega}]$, where $p$ and $q$ denote the position and orientation of the quadrotor, respectively, with their corresponding time derivatives, indicating velocity, represented as $\dot{p}$ and $\omega$. The control space is 4 dimensional with bound $[-15,-5]\times[-1,1]^3$. We randomly place 10 obstacles in the workspace space and ensure the scene is cluttered.

In each of the cases mentioned above, we set up 10 environments by random placement of the obstacles, each with 1000-2000 randomly sampled start and goal state pairs. The $10-20\%$ of data is used for testing, and the remaining for the training. { In our test dataset, we also include two unseen environments for each problem, created by random placement of obstacles. We ensure these environments to be different from the ten seen settings, and for each, we randomly sample 100-200 valid start and goal pairs for evaluation. Hence, in total, our test dataset contains 12 environments for the given setups.} The demonstration trajectories for the training data are obtained using the SST algorithm. Furthermore, we obtain the point-cloud of the environments, by randomly sampling the obstacle space and processing them into voxel $\boldsymbol{v}$ of size $32\times32\times32$.
%We preprocess training dataset and validation dataset in the form of 
%\begin{equation}
%    (O, \boldsymbol{x_t}, \boldsymbol{x_T}; \boldsymbol{x_{t+1}})
%\end{equation}
%where $O$ denotes the voxel representation of the environment; $\boldsymbol{x_t}$, $\boldsymbol{x_{t+1}}$ and $\boldsymbol{x_T}$ denote the current state, the next state and the goal state.
% We found no significant difference between using each waypoint generated by SST, and regularizing the duration between waypoints.

\section{Results} 
%%%%%%%%%%%%%%%%%%%%%
% Trajectory study    
%%%%%%%%%%%%%%%%%%%%%%
\begin{figure*}[ht!]
    %%%%%%%%%%%%%%%%%%%%%%%
    %% Acrobot
    %%%%%%%%%%%%%%%%%%%%%%%
    \begin{subfigure}[b]{0.48\textwidth}
        \includegraphics[width=0.95\textwidth]{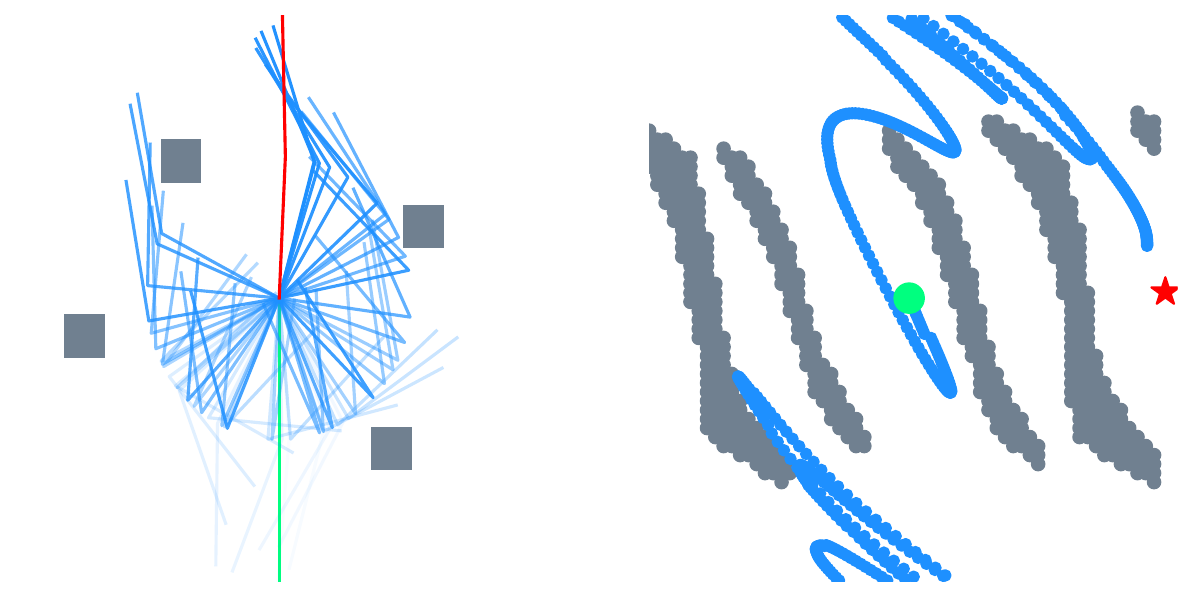}
        \caption{MPC-MPNet: $t=5.5s$, $c=10.9$}
    \end{subfigure}
    \begin{subfigure}[b]{0.48\textwidth}
       \includegraphics[width=0.95\textwidth]{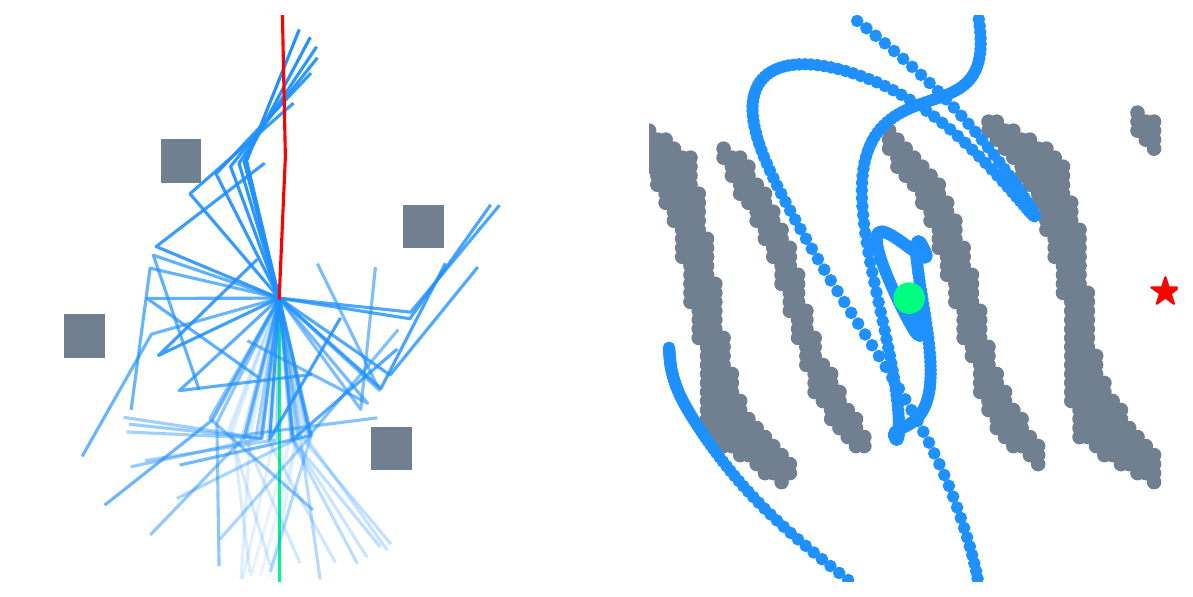}
       \caption{SST: $t=52.4s$, $c=11.2$}
    \end{subfigure}
    \caption{Acrobot environment: The workspace (left) and state-space (right) trajectories are shown in each subfigure. In this example, the start state is  $[0, 0, 0, 0]$, and goal states are randomly distributed around the vertical configuration. From the results, it can be seen that MPC-MPNet could generate a path of comparable lengths as SST with lesser amount of time.}
    \label{fig:acrobot-traj}\vspace*{-0.1in}
\end{figure*}
%%%%%%%%%%%%%%%%%%%%%%
% Trajectory study    
%%%%%%%%%%%%%%%%%%%%%%
We present a set of experiments to compare the computation time, path quality, and success rate of MPC-MPNetPath, MPC-MPNetTree, and SST planning algorithms. %We also experimented with Kinodynamic RRTstar augmented with a BVP solver \cite{xie2015toward}, but due to its significantly larger computation times than all presented methods, we exclude them from the comparison studies. 
All experiments were performed on the same system with 32GB RAM, GeForce GTX 1080 GPU, and 3.40GHz$\times$8 Intel Core i7 processor.

\subsection{Comparative Studies}
This study compares the computation time, success rate, and path quality (measured by time-to-reach) of MPC-MPNetPath, MPC-MPNetTree, and SST algorithms in Cart-Pole, Acrobot, Car, and Quadrotor environments. All evaluation planning problems were unique and not seen during the training, presenting non-trivial and cluttered environments with underactuated systems.

\begin{figure*}[ht!]
    %%%%%%%%%%%%%%%%%%%%%%%
    %% Cart-Pole
    %%%%%%%%%%%%%%%%%%%%%%%
    \begin{subfigure}[b]{0.49\textwidth}
        \raisebox{-.5\height}{\includegraphics[width=0.65\textwidth]{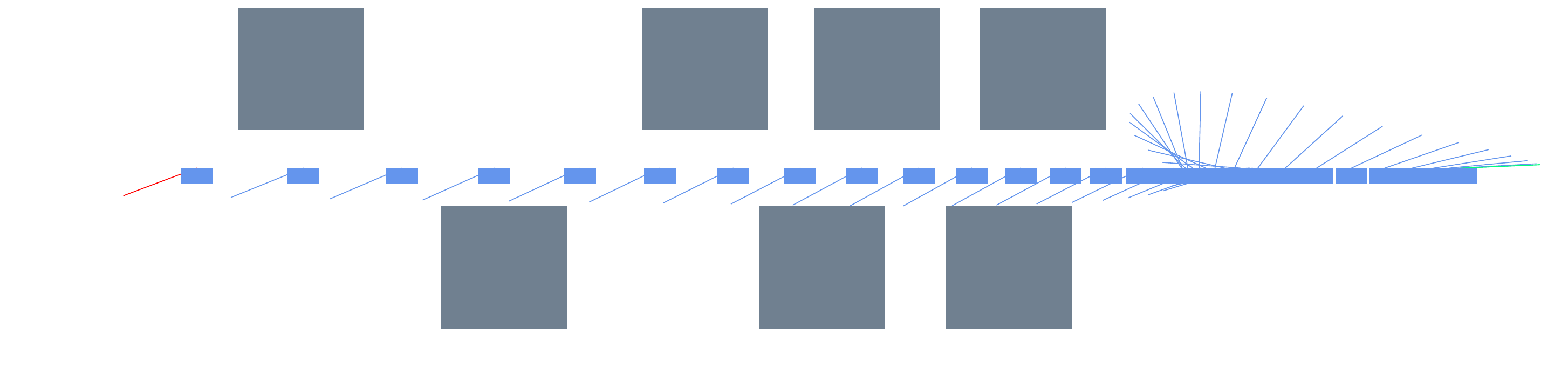}}
        \raisebox{-.5\height}{\includegraphics[width=0.34\textwidth]{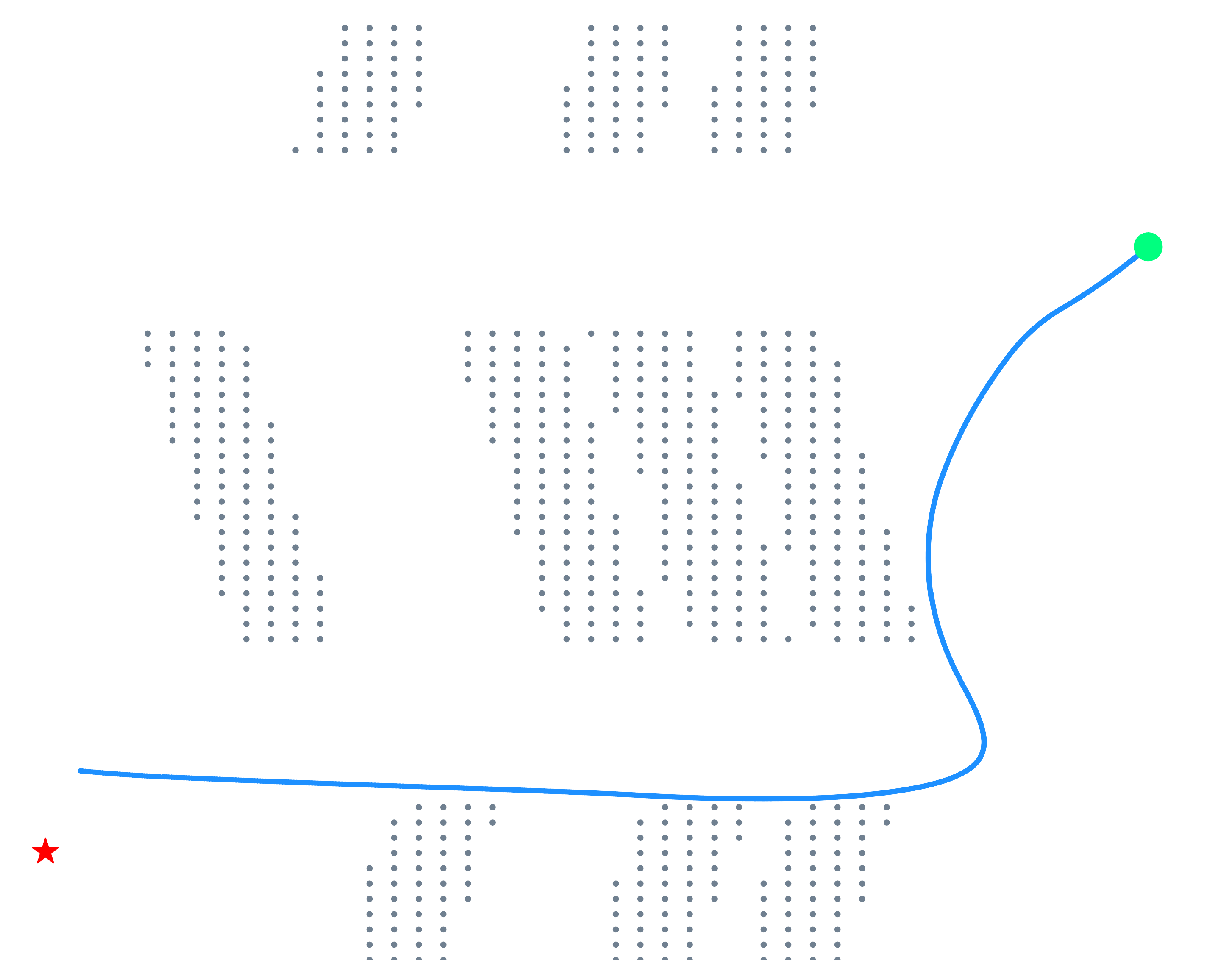}}
        \caption{MPC-MPNet: $t=2.8s$, $c=4.2$}
    \end{subfigure}
    \begin{subfigure}[b]{0.49\textwidth}
        \raisebox{-.5\height}{\includegraphics[width=0.65\textwidth]{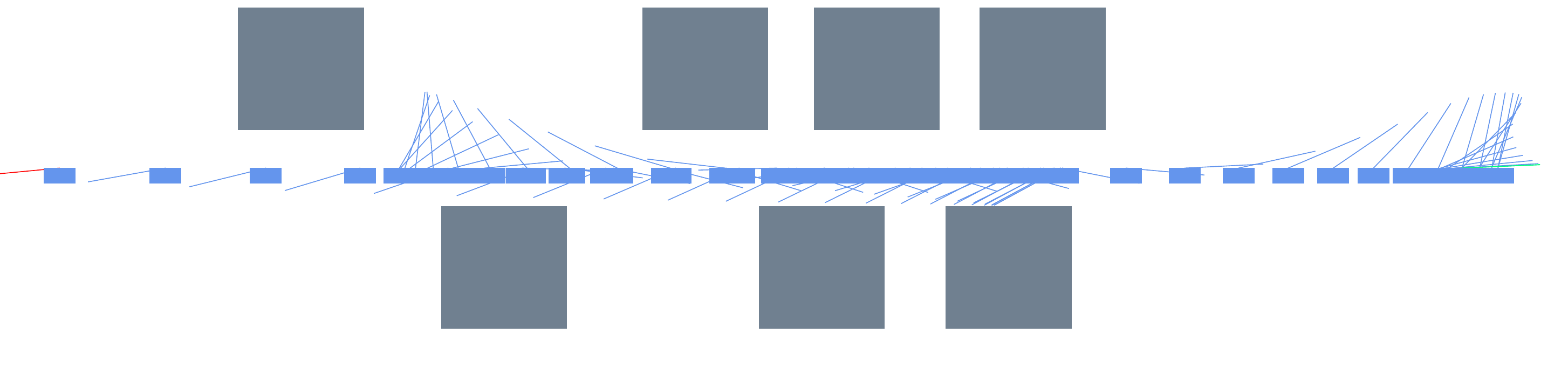}}
        \raisebox{-.5\height}{\includegraphics[width=0.34\textwidth]{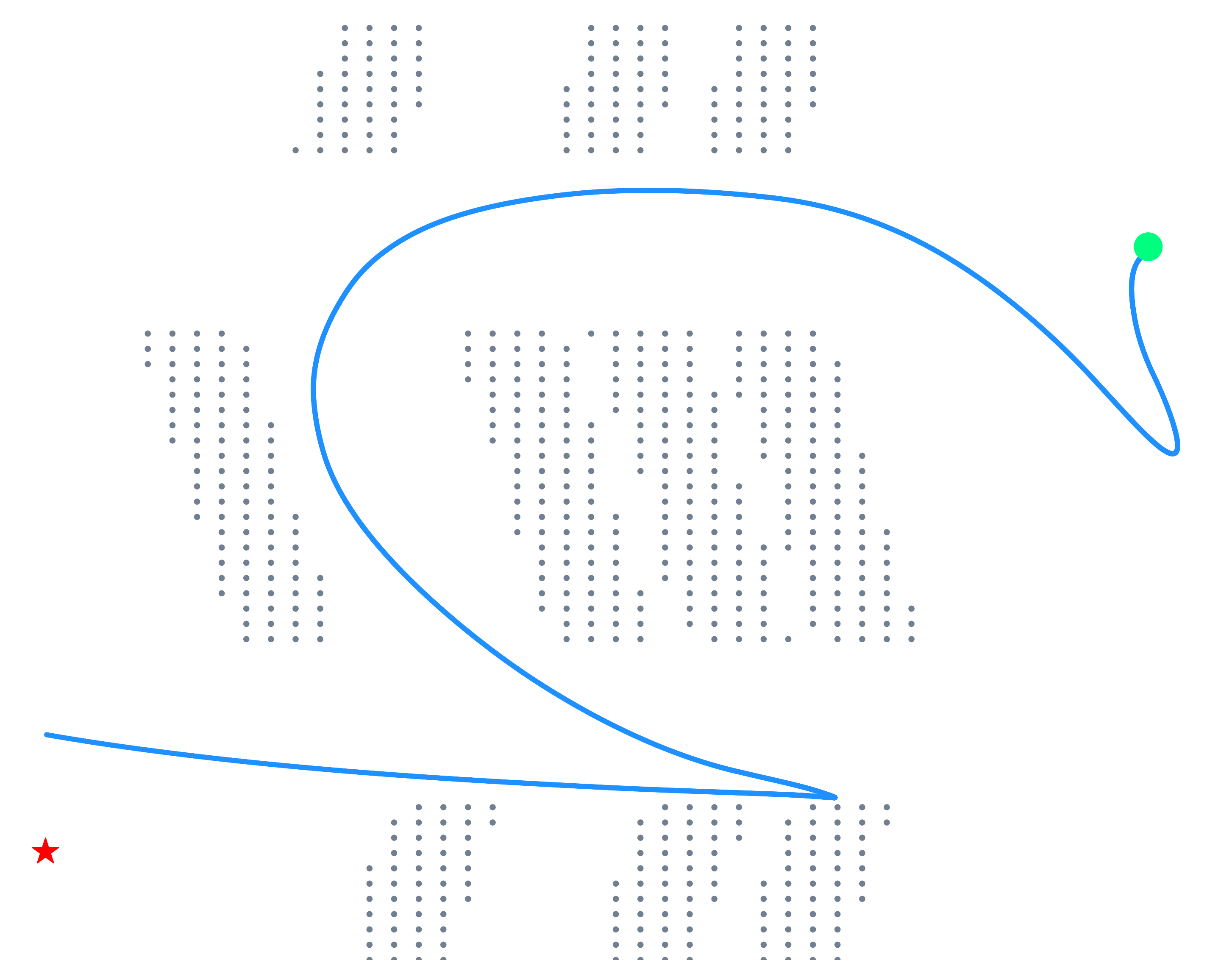}}
        \caption{SST: $t=29.4s$, $c=7.9$}
    \end{subfigure}
    \caption{Cart-Pole environment: The workspace (left) and state-space (right) trajectories are shown in each subfigure. It can be seen that our method shows significant improvement in computation time and path quality over SST in these cluttered scenarios.
}    \label{fig:cartpole-traj}\vspace*{-0.2in}
\end{figure*}

\begin{figure}[ht!]
    %%%%%%%%%%%%%%%%%%%%%%%
    %% Car
    %%%%%%%%%%%%%%%%%%%%%%%
    \begin{subfigure}[b]{0.24\textwidth}
        \includegraphics[width=.95\textwidth]{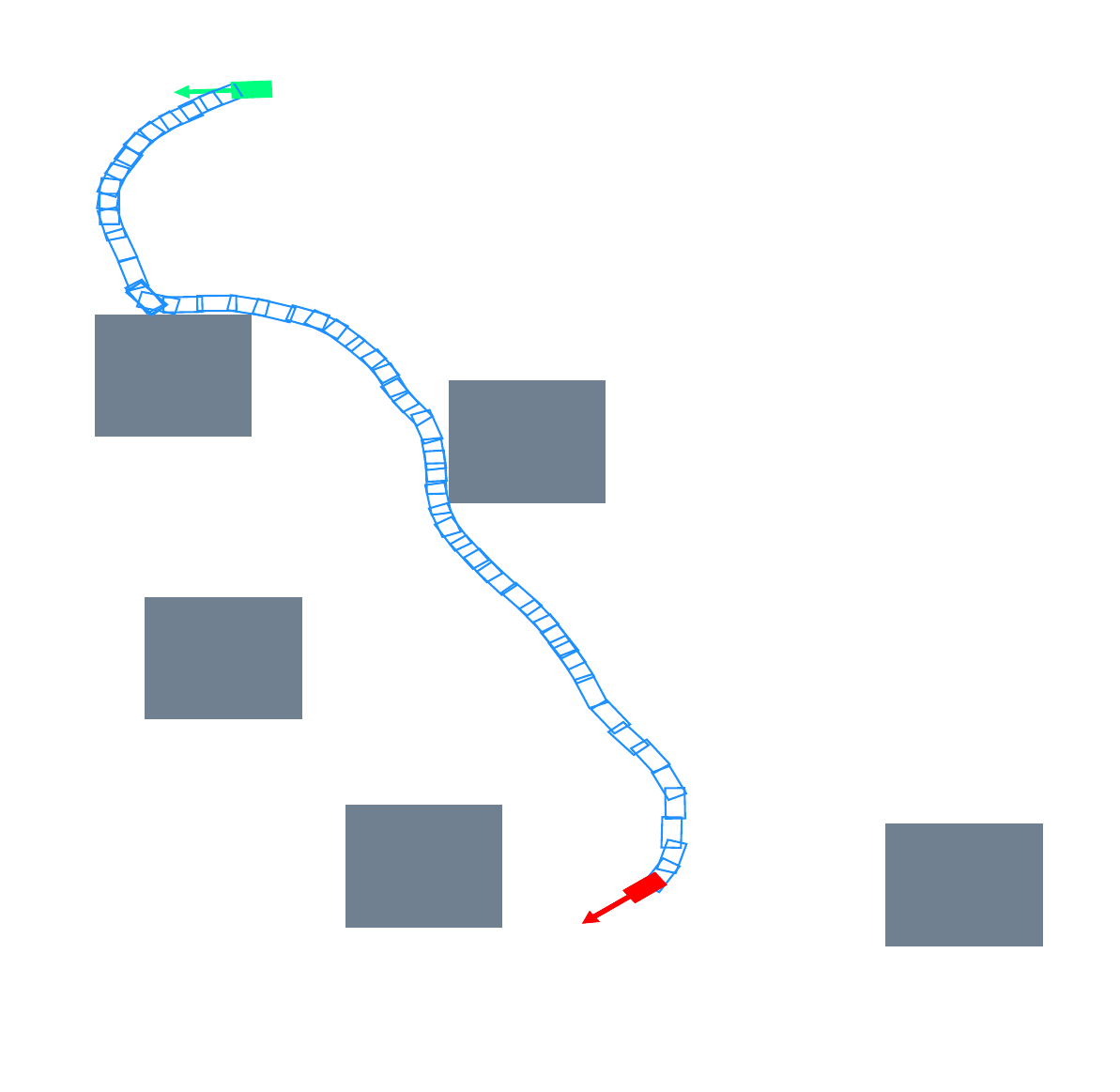}
        \caption{Ours: $t=12.3s$, $c=57.0$}
    \end{subfigure}
    \begin{subfigure}[b]{0.24\textwidth}
        \includegraphics[width=.95\textwidth]{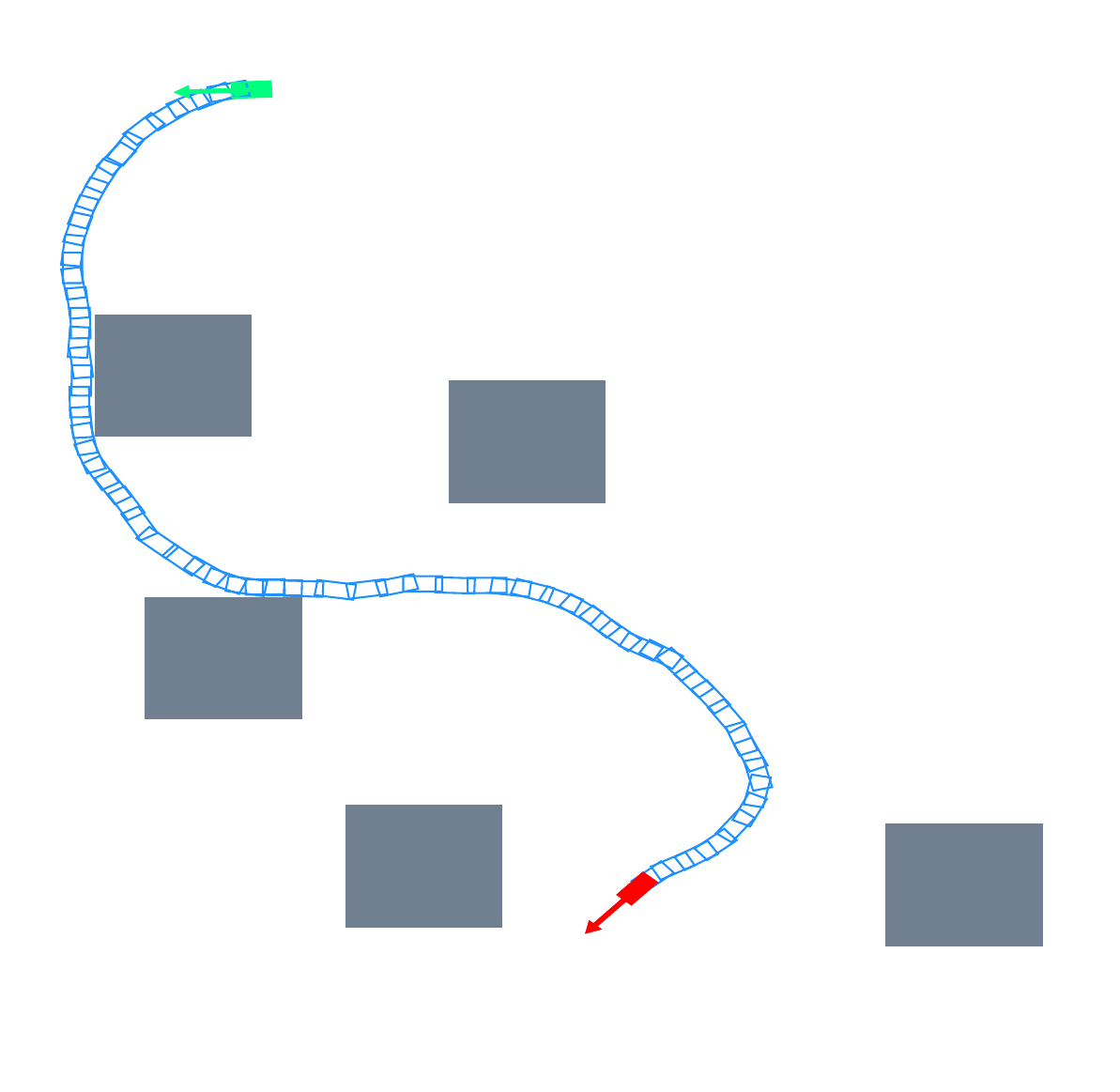}
        \caption{SST: $t=27.2s$, $c=63.8$}
    \end{subfigure}
    \caption{Car environment: MPC-MPNet and SST finding paths under kinodynamic constraints for a non-holonomic system in an example environment with multiple narrow passages.}
    \label{fig:car-traj}
\end{figure}

\begin{figure}[ht!]
%%%%%%%%%%%%%%%%%%%%%%%
%% Quadrotor
%%%%%%%%%%%%%%%%%%%%%%%
    \begin{subfigure}[b]{0.24\textwidth}
        \includegraphics[width=1.1\textwidth]{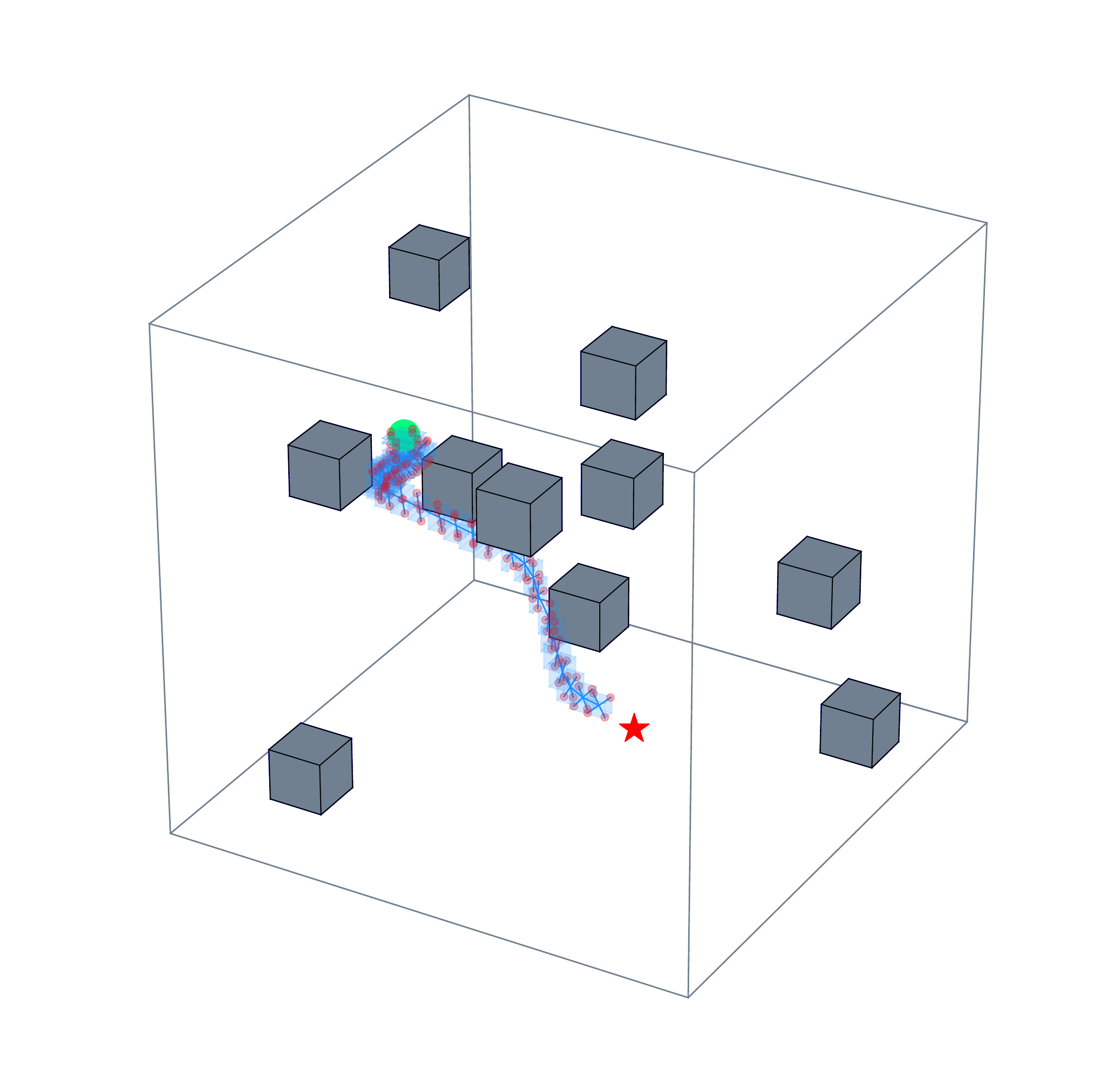}
        \caption{Ours: $t=1.5s$\\$c=7.6$}
    \end{subfigure}
    \begin{subfigure}[b]{0.24\textwidth}
        \includegraphics[width=1.1\textwidth]{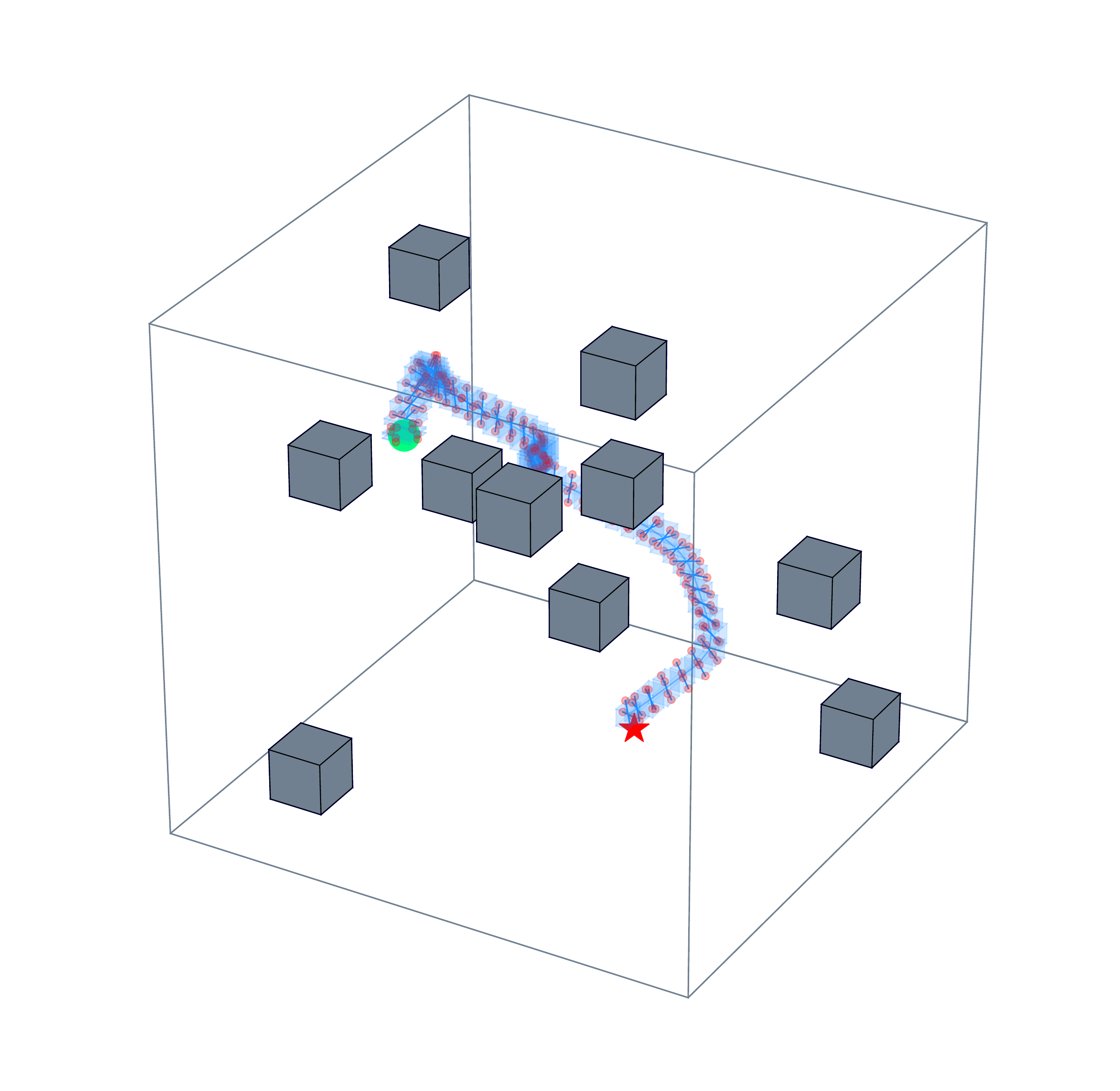} 
        \caption{SST: $t=162.3s$\\$c=9.9$}
    \end{subfigure}
    \caption{Quadrotor environment: The problem requires finding a kinodynamically constrained motion of a 12 DOF quadrotor in challenging environments. In these scenarios, our methods were at least 50 times faster than SST, showing our method's scalability to high-dimensional KMP problems.}
    \label{fig:quadrotor-traj}
    \vspace{-5mm}
\end{figure}

{Table~\ref{tab:results} presents the mean computation times with standard deviations across different scenarios in both seen and unseen environments. Figs.~\ref{fig:boxplot-acrobot-cartpole-car} show the box-plots of all methods comparing their computation time and path quality inter-quartile ranges for solving all kinodynamic planning problems. Fig.~\ref{fig:acrobot-traj} to Fig.~\ref{fig:quadrotor-traj} show example qualitative results from MPC-MPNet and SST. In all these problems, including unseen scenarios, MPC-MPNetTree and MPC-MPNetPath success rates were between $90-100\%$ and $85-95\%$, respectively, comparable to the SST success rates in the given time limit.}

It can be seen that compared to SST, MPC-MPNet methods take significantly lower computation times to find similar quality path solutions with comparable success rates. Our experiments also show that for environments with high-dimensional state and control spaces such as in Quadrotor, SST's computation times increase exponentially. In contrast, MPC-MPNet still retains its computational benefits from informative waypoint sampling and outperforms SST by a large margin. 

Among MPC-MPNetTree and MPC-MPNetPath, we observed that the former method achieves higher success rates and finds better quality path solutions by exploiting GPU-accelerated, parallel computation. Nevertheless, we present CPU-based MPC-MPNetPath and GPU-based MPC-MPNetTree to highlight that both approaches perform better than traditional gold-standard planning methods. Moreover, our algorithms balance exploration and exploitation through randomly sampling current configurations within trees to provide better completeness and comparable success rates as conventional methods.
%%%%%%%%%%%%%%%%%%%%%%%
% Ablation study table 
%%%%%%%%%%%%%%%%%%%%%%%
\begin{table*}[ht!]
\centering 
\begin{tabular}{ccccc}\toprule
\multirow{2}{*}{Setup}&\multicolumn{4}{c}{Planning Tasks}\\\cmidrule{2-5}
&\multicolumn{1}{c}{Acrobot}&\multicolumn{1}{c}{Cart-Pole}&\multicolumn{1}{c}{Quadrotor}&\multicolumn{1}{c}{Car-like}\\\midrule
w/o D & 
    $8.14 \pm 10.37 (7.23 \pm 4.91)$ & 
    $5.43 \pm 8.59 (7.67 \pm 3.37)$ & 
    $0.81 \pm 4.11 (8.45 \pm 4.11)$ & 
    $13.29 \pm 13.36 (47.72 \pm 37.36)$ \\
w/ D & 
    $\boldsymbol{5.09 \pm 6.87} (\boldsymbol{5.18 \pm 3.82})$ &
    $\boldsymbol{4.47 \pm 7.37} (\boldsymbol{6.25 \pm 3.13})$ &  
    $\boldsymbol{0.46 \pm 2.12} (\boldsymbol{6.49 \pm 2.71})$ & 
    $\boldsymbol{8.96 \pm 12.33} (\boldsymbol{47.68 \pm 20.44})$ \\
\bottomrule
\end{tabular}
\caption{ Ablation Study: The total mean computation time with and with out neural discriminator is shown for MPC-MPNetPath, where the path quality, measured by the time-to-reach, is presented in parentheses.}
\label{tab:ablation-for-DNet}
\end{table*}
\subsection{Ablation Studies}
%%%%%%%%%%%%%%%%%%%%%%%
% Ablation study
%%%%%%%%%%%%%%%%%%%%%%%

To show the impact of the neural discriminator in the planning pipeline, we present an ablation study, comparing the planning time and path quality between (i) MPC-MPNetPath without neural discriminator, which predicts only one state at every iteration, and (ii) MPC-MPnetPath, which generates a batch of waypoints with the neural generator, and selects the best based on an estimated cost by the neural discriminator. All experiments are conducted in the same environmental setup as in the comparative studies.

From the results shown in Table~\ref{tab:ablation-for-DNet}, we can observe that the neural discriminator contributes to reducing the mean and standard deviation of planning time. It also helps with generating trajectories with a better cost quality. In MPC-MPNet, stochasticity is introduced by dropout layers in the neural generator. This operation generates a variety of samples, some of which might require replanning for the connections. Our process eliminates those cases using the neural discriminator by selecting a state with the minimum time to reach the given target, also resulting in fewer planning iterations and better path quality.

\section{Discussion}

In this section, we highlight the worst-case properties of our
proposed algorithms named MPC-MPNetPath and MPC-MPNetTree. The former operates MPC without GPU processing and uses discriminator to remove unnecessary states to save computational resources. The latter builds on GPU programming and parallelly computes various nodes for simultaneous local extensions with MPC. Our methods plan without relying on bidirectional tree generation or any replanning as was required in the original MPNet and its variants. However, similar to MPNet, our neural generator does explore the sub-space of given state-space that potentially contains a path solution, thanks to Dropout, that implants randomness into our neural generator adapted from expert demonstrations. Since our generated state-spaces are confined to a subspace, the resulting informed tree also grows in that region due to directed extensions with MPC. To enable our methods to explore the state-space outside the generator's learned state-spaces, we propose a notion of stage-wise exploration.

Our stage-wise exploration strategy balances global exploitation-exploration in three phases. In the first phase, our proposed algorithms are operated for a fixed number of planning iterations $N_1$. In the second phase, i.e., after $N_1$ iterations and until $N_2$ iterations, our algorithm replaces the MPC-based local controller with a random shooting method. This phase allows exploration in the action-spaces while still keeping state-spaces informed as given by the neural generator. Finally, after $N_2$ steps and beyond in the third phase, our approach performs full exploration by randomly sampling both state and action spaces, like the SST algorithm.  These three phases allow our trees to expand from an inner region, potentially containing a path solution, to an outer region in the worst-case for finding a solution if one exists. In our experimentation, we validated that, similar to MPNet, incorporating staged-wise exploration ensures $100\%$ success rate while still retaining the computational benefits and performing better than classical planning approaches. 

Since our methods perform stage-wise exploration of the state and action spaces and eventually explore them entirely over a large number of planning iterations, the resulting approach exhibits \textit{probabilistic-completeness}. It implies that MPC-MPNet finds a path, if one exists, with the probability approaching to one as the number of planning iterations reaches infinity. The formal proofs can be derived in the same way as reported in \cite{li2016asymptotically}. Furthermore, note that our MPC-MPNetTree method also uses the nearest neighbor search for extending trees. This method is adopted from SST, which selects the best neighbors in terms of their cost from the given start/root state and removes tree edges with relatively higher costs. Based on this nearest-neighbor selector and the random exploration, the SST method \cite{li2016asymptotically} also guarantees \textit{asymptotic-optimality}, i.e., over a large number of planning iterations, their planner will eventually find a minimum cost path solution, if one exists. Since MPC-MPNetTree also does exploration, though in stages, and uses SST's like nearest node selector and pruner, it also exhibits \textit{asymptotic-optimality} with proofs being similar to as presented in \cite{li2016asymptotically}.    

\section{Conclusions}
We propose Model Predictive Motion Planning Network or MPC-MPNet, a learning-based algorithm capable of finding solutions in seconds with near-optimal path quality in complex kindynamically constrained scenarios where other state-of-the-art methods take upto minutes for finding a similar quality path solutions. Besides, our experiments show that MPC-MPNet generalizes to unseen tasks and adapts to high-dimensional problems with high success rates. We also show that MPC-MPNet allows parallel computing and exhibits worst-case theoretical completeness and optimality guarantees, making it an ideal planner for practical problems ranging from robot navigation to full-body humanoid motion planning.  In our future research direction, we would like to investigate our models' generalization to new planning problem domains. We are also interested in extending our approach to planning under dynamic and manifold kinematic constraints for robot manipulation problems.
% In our future studies, we plan to extend MPC-MPNet to high-dimensional
% \clearpage

%%%%%%%%%%%%%%%%%%%%%%%%%%%%%%%%%%%%%%%%%%%%%%%%%%%%%%%%%%%%%%%%%%%%%%%%%%%%%%%%

%%%%%%%%%%%%%%%%%%%%%%%%%%%%%%%%%%%%%%%%%%%%%%%%%%%%%%%%%%%%%%%%%%%%%%%%%%%%%%%%

%%%%%%%%%%%%%%%%%%%%%%%%%%%%%%%%%%%%%%%%%%%%%%%%%%%%%%%%%%%%%%%%%%%%%%%%%%%%%%%%

\bibliographystyle{IEEEtran}
\bibliography{ref}

\end{document}